\documentclass[conference]{IEEEtran}
\IEEEoverridecommandlockouts
\usepackage{cite}
\usepackage{amsmath,amssymb,amsfonts}
\usepackage{algorithmic}
\usepackage{graphicx}
\usepackage{textcomp}
\usepackage{xcolor}
\def\BibTeX{{\rm B\kern-.05em{\sc i\kern-.025em b}\kern-.08em
    T\kern-.1667em\lower.7ex\hbox{E}\kern-.125emX}}
\begin{document}

\title{Investigating Continuous Learning in Spiking Neural Networks 
}

\author{\IEEEauthorblockN{ C. Tanner Fredieu}
\IEEEauthorblockA{\textit{Sanghani Center for Artificial Intelligence and Data Analytics} \\
\textit{Bradley Department of Electrical and Computer Engineering, 
Virginia Tech, Blacksburg, VA, USA} 
\\Email: christianf@vt.edu
}}



\maketitle

\begin{abstract}
In this paper, the use of third-generation machine learning, also known as spiking neural network architecture, for continuous learning was investigated and compared to conventional models. The experimentation was divided into three separate phases. The first phase focused on training the conventional models via transfer learning. The second phase trains a Nengo model from their library. Lastly, each conventional model is converted into a spiking neural network and trained. Initial results from phase 1 are inline with known knowledge about continuous learning within current machine learning literature. All models were able to correctly identify the current classes, but they would immediately see a sharp performance drop in previous classes due to catastrophic forgetting. However, the SNN models were able to retain some information about previous classes. Although many of the previous classes were still identified as the current trained classes, the output probabilities showed a higher than normal value to the actual class. This indicates that the SNN models do have potential to overcome catastrophic forgetting but much work is still needed.
\end{abstract}


\section{Introduction}
\subsection{Continuous Learning in Artificial and Biological Systems}
Continuous learning systems are predicated on the notion that natural learning occurs overtime and when new information enters an environment. In biological organisms, this is clearly indicated with how they naturally adapt to their environments even when new challenges arise either through a form of supervised, unsupervised, or reinforcement learning strategies.

The integration of continuous learning into artificial intelligence is not a new endeavor, but it is one that has alluded researchers since the beginning. The biggest challenge is overcoming catastrophic forgetting [3]. Catastrophic forgetting occurs when prior information about previous classes is lost when learning information about new classes. Many different methods have been tried in attempts to solve the problem. Some of these include new architecture designs [17], new learning strategies such as integrating data from previous classes into the new training data [13] as a form of refresher, and possible replacement of backpropagation with more bio-inspired methods [9]. One of the most promising routes is to use artificial neural networks and hardware that more closely mimic the brain such as with spiking neural networks [1] and neuromorphic hardware [4].


\subsection{Spiking Neural Networks}
Spiking neural networks (SNNs) are a type of neural network that seeks to mimic the transfer of information inside biological neurons [1]. This is accomplished by using a temporal dimension to identify the timing or spiking of each neuron. The sequence of the firing neurons are known as a spike train [1]. Because of this, SNNs are considered to be the basis for third-generation artificial neural networks.

One of the largest advantages to SNN models are their low-power consumption and efficiency [7] when compared to conventional models especially when paired with neuromorphic hardware which will be discussed in the next section.

Currently, training SNNs proves to be a challenge due to a number of hurdles. First, training SNNs conventionally is difficult due to the non-differentable nature of the neurons [9]. There have been a number of ways to overcome this such as converting conventional models to SNNs [12] which is what was performed in this experimentation. However, this comes at the cost of increasing inference latency and losing information during conversion. Another hurdle is the time and complexity needed when training these networks. As stated previously, neuromorphic hardware will greatly increase the use of SNNs, but they are also more of a necessity than complimentary. Training on current hardware proves to be difficult due to the complexity of the spikes from the neurons [9].

\subsection{Neuromorphic Computing}
While SNNs provide the software portion of the third-generation, a hardware solution must also accompany it much like graphic processing units (GPUs) did for deep learning in the 2010s. This type of new, specialized hardware is known as neuromorphic computing. Much like SNNs, neuromorphic computing strives to mimic the physical structuce and benefits of the brain [4]. This includes faster processing times and lower power consumption [4]. These new hardware platforms also promise to allow for greater complexity where SNNs can make the greatest use. This will solve the previously mentioned  problem of training SNNs as training on current hardware is time consuming due to the nature of SNNs being different than those of conventional models [12]. While much research is being devoted to neuromorphic computing both industrially and academically [4,8,16], commercial use of this hardware is not yet achievable due to a number of reasons.

\begin{figure}[t]
\centerline{\includegraphics[scale=0.48]{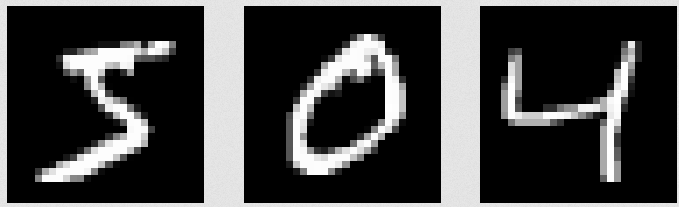}}
\caption{MNIST Samples}
\label{fig3}
\end{figure}

\begin{table}[t]
\caption{Performance across Pre-trained models MNIST}
\begin{center}
\begin{tabular}{|c|c|c|c|}
\hline
\cline{2-4} 
\textbf{Model} & \textbf{\textit{1st MNIST}}& \textbf{\textit{3rd MNIST}}& \textbf{\textit{5th MNIST}} \\
\hline
ResNet50 & 99.12 & 32.33 & 19.70 \\
\hline
ResNet101 & 99.54 & 32.40 & 19.65 \\
\hline
VGG19 & 98.73 & 30.14 & 18.95 \\
\hline
\end{tabular}
\label{tab1}
\end{center}
\end{table}

\section{Experimental Setup}
The experimentation was divided into three different phases. In phase 1, conventional models were trained and tested using incremental learning strategies to observe their limitations and benefits. All models of this type were pre-trained models that used transfer learning from ImageNet weights. In total three different pre-trained models were used, ResNet50, ResNet101 and VGG19. Phase 2 used the original model developed by Nengo for their tutorial on converting conventional models to SNN models. Lastly, phase 3 used the NengoDL library to convert the conventional models from phase 1 to SNNs to investigate the impact of transfer learning. 

Training and testing were kept as minimalistic as possible due to time and resource constraints. All experimentations were performed on Google Colaboratory using TensorFlow 2.11, Python 3, and the Nengo API which will be discussed more in the next section. Training was conducted by increments of two classes. As each dataset contains 10 classes, each model was trained a total of 10 separate times to test the ability to remember previous information. This means that at each increment any previous class training data would not be present within the new training data. 5 total training phases are used for MNIST training data, and 5 total training phases are used for the Fashion MNIST training data.

\subsection{Nengo API and Software}

To implement SNNs as well as convert conventional models to SNNs, a software package is needed to perform the tasks efficiently and quickly rather than from scratch each time. This is similar to how machine learning researchers now make use of frameworks such as Pytorch and TensorFlow to rapidly prototype models. There are many openly available packages online such as PySNN, snnTorch, and Nengo. For this experimentation, Nengo was chosen for converting and training the SNN models.

Nengo created an API that allows for ease of use for creating conventional models inside the TensorFlow framework then converting them efficiently to SNN model equivalents. This ease of use along with documentation sources are the primary reasons it was chosen for the experimentation. One of the major goals of Nengo is create a framework that is ubiquitous for all neuromorphic computing hardware. 

\subsection{Datasets}
As stated earlier, the common benchmark datasets of MNIST and Fashion MNIST were used to provide consistent comparisons across all models. Each dataset contains 10 classes with 60,000 training and 10,000 test images of size 28x28. A sample of the digits contained in MNIST is illustrated in Figure 1, and a sample of the clothing contained in Fashion MNIST is illustrated in Figure 2.

\subsection{Regularization and Optimization}

Regularization and optimization of the models were kept at a constant due to many of these models already have established optimal hyperparameters. As such, the conventional models in phase 1 used the Adam optimizer [2] with a learning rate of 0.001 and L2 regularization.

Regularization and optimization for the SNNs are different as the goal is to optimize the rate of firing of each neuron along with smoothing the synapses for the regularization. Consideration for the value of the firing rate must also take into account the trade-off for accuracy versus inference time. As the number of firings per neuron increases, so too does the latency of the inference. After various testing over different values, the optimal fire rate for the neurons was between 10 and 20 while the optimal synapses smoothing was found to be 0.001 for the regularization.

\subsection{Training and Testing}

Each model was trained on new classes for 10 epochs with a batch size of 200 for each increment. The total amount of training images used per increment varied due to the difference in number of images per class. However, after all 5 increments, the model had seen all 60,000 training images from the target dataset. The same occurs with the test images until the models were tested on all 10,000 images.


\begin{figure}[t]
\centerline{\includegraphics[scale=0.48]{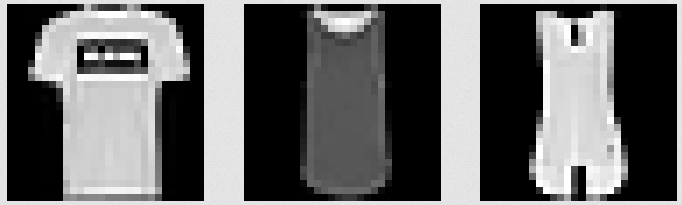}}
\caption{Fashion MNIST Samples}
\label{fig3}
\end{figure}

\begin{table}[t]
\caption{Performance across Pre-trained models F-MNIST}
\begin{center}
\begin{tabular}{|c|c|c|c|}
\hline
\cline{2-4} 
\textbf{Model} & \textbf{\textit{1st F-MNIST}}& \textbf{\textit{3rd F-MNIST}}& \textbf{\textit{5th F-MNIST}} \\
\hline
ResNet50 & 98.92 & 31.32 & 19.54 \\
\hline
ResNet101 & 99.06 & 32.03 & 19.60 \\
\hline
VGG19 & 98.07 & 29.66 & 18.88 \\
\hline
\end{tabular}
\label{tab1}
\end{center}
\end{table}

\section{Results}

\subsection{Pre-trained Models}



Table 1 and Table 2 provides the performance data of each pre-trained model during their separate testings for the MNIST and Fashion MNIST datasets. In the first testing increment of the first two classes of each dataset, all models show performances that are previously known from the literature and machine learning community. Catastrophic forgetting becomes obviously apparent in the next few testing increments when new classes are introduced. As new classes are added, the accuracy of the models on previous classes decreases dramatically while current classes see high-performance accuracy. Performance accuracy above 99\% is retained for the new classes while dropping to around 30\% by the third set of training all the way to below 20\% by the time all 10 classes have been trained on by the models.

\subsection{SNNs}

\begin{table}[t]
\caption{Performance across SNN models MNIST}
\begin{center}
\begin{tabular}{|c|c|c|c|}
\hline
\cline{2-4} 
\textbf{Model} & \textbf{\textit{1st MNIST}}& \textbf{\textit{3rd MNIST}}& \textbf{\textit{5th MNIST}} \\
\hline
SNN-ResNet50 & 99.87 & 40.17 & 30.55 \\
\hline
SNN-ResNet101 & 99.88 & 40.54 & 30.78 \\
\hline
SNN-VGG19 & 99.67 & 39.98 & 29.56 \\
\hline
SNN & 99.45 & 39.97 & 29.44 \\
\hline
\end{tabular}
\label{tab1}
\end{center}
\end{table}

\begin{table}[t]
\caption{Performance across SNN models F-MNIST}
\begin{center}
\begin{tabular}{|c|c|c|c|}
\hline
\cline{2-4} 
\textbf{Model} & \textbf{\textit{1st F-MNIST}}& \textbf{\textit{3rd F-MNIST}}& \textbf{\textit{5th F-MNIST}} \\
\hline
SNN-ResNet50 & 99.66 & 39.95 & 30.02 \\
\hline
SNN-ResNet101 & 99.75 & 40.15 & 30.18 \\
\hline
SNN-VGG19 & 99.38 & 39.24 & 29.09 \\
\hline
SNN & 99.05 & 39.33 & 29.13 \\
\hline
\end{tabular}
\label{tab1}
\end{center}
\end{table}

 In this section, the results from the testing of the different SNN models are observed and discussed. Like the previous section the performance results of each model during testing are illustrated in Table 3 and Table 4. The performances of all the models follow a similar trend as those observed with the conventional models. With each new set of classes add, the accuracy of those classes are high while the accuracy of previous classes decreases. However, unlike previous examples, the information about previous classes seems to be somewhat retained.

\begin{figure}[t]
\centerline{\includegraphics[scale=0.4]{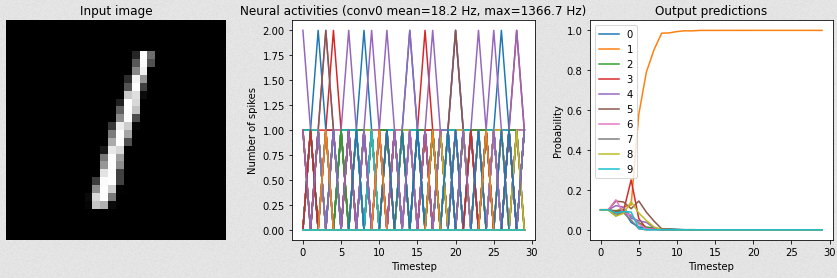}}
\caption{Accuracy of sample from current MNIST class}
\label{fig3}
\end{figure}

Figure 3, Figure 6, and Figure 7 gives confirmation of the models ability to learn the current classes effectively from the MNIST and Fashion MNIST datasets. The first frame illustrates the image that is being classified while the second frame shows which neurons are spiking in the first convolutional layer after the input layer. Finally, the output prediction is shown in the final frame.

Figure 8 shows a different classification of the image from Figure 6 after the model has gone through incremental training on new classes. This clearly illustrates the model has experienced a form of catastrophic forgetting as the model attempts to classify the image as one of the new classes. This is also evident in the changes in neuron spiking in the second frame when compared to the same frame from Figure 6. However, it is also illustrated that the correct class still retains a significant output probability even though catastrophic forgetting has occurred. 

This same result is shown again in Figure 4 and Figure 5. Though, Figure 4 shows just how little information can be retained at times. This would explain why the SNN models are able to slightly outperform the conventional models but to a limited degree. The primary drawback of the SNN models, as discussed previous from [1], becomes evident from these observations. With each variation in firing rate to regularize the model as well as when new classes are added, the latency of inference becomes larger as illustrated in the second frame of the figures.

\begin{figure}[t]
\centerline{\includegraphics[scale=0.4]{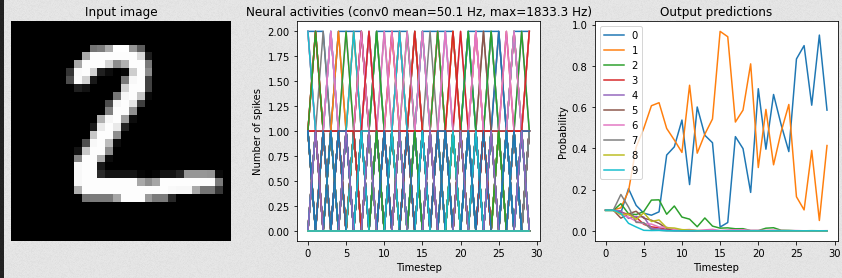}}
\caption{Accuracy of sample from previous MNIST class}
\label{fig3}
\end{figure}

\begin{figure}[b]
\centerline{\includegraphics[scale=0.4]{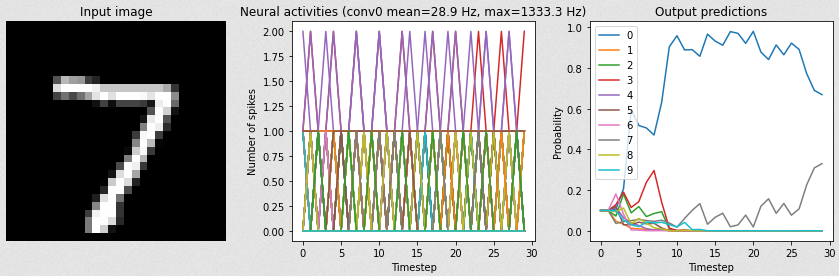}}
\caption{Information retention of previous MNIST class}
\label{fig3}
\end{figure}

\begin{figure}[t]
\centerline{\includegraphics[scale=0.4]{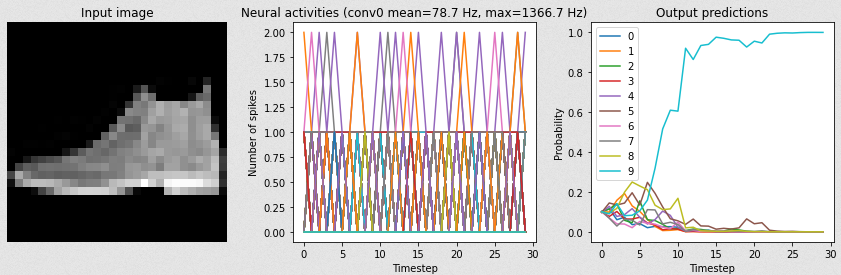}}
\caption{Accuracy of sample from current (1st) F-MNIST class}
\label{fig3}
\end{figure}

\begin{figure}[t]
\centerline{\includegraphics[scale=0.4]{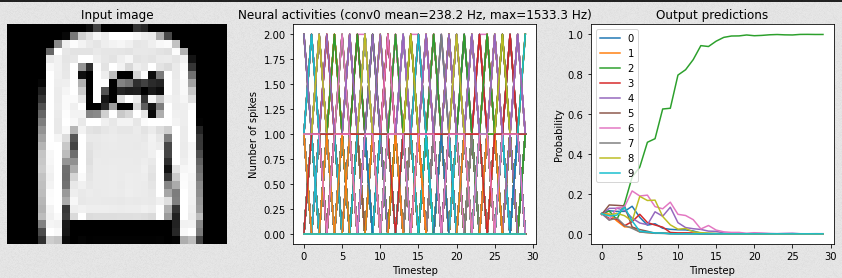}}
\caption{Accuracy of sample from current (2nd) F-MNIST class}
\label{fig3}
\end{figure}


\break

\section{Future Directions}
For future endeavors, there are a few different routes that can be investigated. The first is trying to solve the problem occurring with converted models. Information about previous classes are shown to be somewhat retained. Although, this isn't enough to get an accurate classification. A possible solution may also lie within biological processes. The role of dreaming in the artificial sense might be able to strengthen the old information. Dreaming is known to be necessary in biological organisms to allow for successful organization of new and old information [5]. It has also been shown that computer architecture [5] and reinforcement learning environments that mimic dreaming are able to performed better in old and even new environments that are similar [11].

The second route is far less clear. Some researchers [9,14,15] have suggested a whole new approach to artificial intelligence learning that is closer to biological processes such as how the brain uses plasticity. This would mean that the current success of conventional models and backpropagation may not be able to be used such as with the methods used in this experimentation. Both should be investigated thoroughly.

The third and most immediate route is to experiment with other successful architectures such as vision transformers [10] and other learning strategies such as unsupervised and reinforcement learning.

\section{Conclusion}
In conclusion, the experimentation showed that there is potential in the use of spiking neural networks for the realization of continuous learning. However, there is much work to be accomplished in this area. It is shown that after incremental training the SNNs are able to retain some level of information from the previous classes when compared to conventional models, but they still are unable retain enough information to correctly identify them from the current classes used for training. The next step is to investigate methods that may be able to help improve information retention such as a way to mimic the organization of knowledge during the dreaming stage in biological organisms or perhaps a new learning strategy overall.

\begin{figure}[t]
\centerline{\includegraphics[scale=0.4]{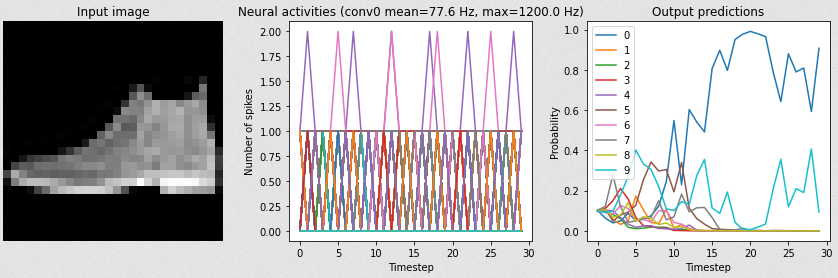}}
\caption{Information retention of previous F-MNIST class}
\label{fig3}
\end{figure}

\vspace{12pt}

\end{document}